# Characterizing the Weight Space for Different Learning Models

Saurav Musunuru
Computer Science & Engineering
Indian Institute of Technology Delhi, India
email: musunurusaurav@gmail.com

Jay N. Paranjape
Computer Science & Engineering
Indian Institute of Technology Delhi, India
email: jay.edutech@gmail.com

Rahul Kumar Dubey
Robert Bosch
Engineering and Business Solutions
Bangalore, India
email: RahulKumar.Dubey@in.bosch.com

Vijendran G. Venkoparao
Robert Bosch
Engineering and Business Solutions
Bangalore, India
e-mail: GopalanVijendran.Venkoparao@in.bosch.com

*Abstract*—Deep Learning has become one of the primary research areas in developing intelligent machines. Most of the well-known applications (such as Speech Recognition, Image Processing and NLP) of AI are driven by Deep Learning. Deep Learning algorithms mimic human brain using artificial neural networks and progressively learn to accurately solve a given problem. But there are significant challenges in Deep Learning systems. There have been many attempts to make deep learning models imitate the biological neural network. However, many deep learning models have performed poorly in the presence of adversarial examples. Poor performance in adversarial examples leads to adversarial attacks and in turn leads to safety and security in most of the applications. In this paper we make an attempt to characterize the solution space of a deep neural network in terms of three different subsets viz. weights belonging to exact trained patterns, weights belonging to generalized pattern set and weights belonging to adversarial pattern sets. We attempt to characterize the solution space with two seemingly different learning paradigms viz. the Deep Neural Networks and the Dense Associative Memory Model, which try to achieve learning via quite different mechanisms. We also show that adversarial attacks are generally less successful against Associative Memory Models than Deep Neural Networks.

## I. Introduction

Today, Deep Learning has become one of the primary research areas in developing intelligent machines. Most of the well-known applications of AI, like Image Classification, Recognition, and Computer Vision and so on [1] [2] are driven by Deep Learning. Today, it is being used in many real world applications like Automated Driving[3], Anomaly Detection[4] and Healthcare[5]. Deep Learning algorithms mimic human brains using artificial neural networks and progressively learn to accurately solve a given problem. But, with great technological advances come complex difficulties and hurdles. For the reliable application of Deep Neural Networks in the domain of security, the robustness against adversarial attacks must be well established. There had been several speculative explanation regarding the existence of adversarial examples. Some of the explanations attribute this to the non-linearity of deep neural networks, but recently in [6], the authors showed that linear behavior in high dimensional spaces is sufficient to produce adversarial examples in neural networks. In this paper, we study and compare two such models namely the Deep Neural Networks and the Associative Memory, which try to achieve learning via quite different mechanisms.

### A. Deep Neural Networks(DNN)

Supervised learning is a glorified name for a function which takes some input and gives an output according to the application it is used for. For example, in image classification tasks, an image is given as an input and the function outputs a set of numbers, the maximum of which is the predicted label of the image. This is but one technique of classification and the task of classification is but one among many, the general idea behind neural networks in a supervised learning setup remains the same i.e. find a mapping from input to output.

What makes neural networks so useful is the fact that these mappings don't have to be identified by the programmer. Instead, the programmer defines an architecture of layers performing simple mathematical functions like arithmetic, matrix multiplication, convolution, non-linear functions like tanh, and so on, and also provides this model with a large set of input-output pairs. Through multiple passes of the input, and update rules defined by the algorithms such as Statistical Gradient Descent (SGD), the model 'learns' a set of weights which it uses to perform the mathematical operations so that maximum of the given pairs are satisfied.

When many such simple mathematical layers are used between the input and the output layers, the network is termed as 'Deep' Neural Network. These models have performed extremely well in almost all kinds of machine learning tasks. Well known examples include Oxford's VGG16[2], Google's GoogleNet[7] and so on.

### B. Associative Memory

An associative memory is a type of Neural Network paradigm that allows for the recall of data based on the degree of similarity between the input pattern and the patterns stored in memory. It refers to a memory organization in which the memory is accessed by its content as opposed to an explicit address like in the traditional computer memory system. Therefore, this type of memory allows the recall of information based on partial knowledge of its contents. Hopfield Network was first introduced by J.J.Hopfield in 1982[8]. It's structure is based on the biological neuron. Similar to the human brain, the Hopfield network memorizes the given patterns. And in the process of recall it tries to recall the memorized pattern which is closest to the given pattern. In this sense the Hopfield Network functions as an associative memory. There are multiple learning rule for Hopfield networks. The training of the Hopfield network is finding the weight matrix for the weights which connects the neurons. The Hebbian learning rule is

$$W_{ij} = \frac{1}{N} \sum_{\mu=1}^{P} \xi_i^\mu \cdot \xi_j^\mu \quad (1)$$

We use an iterative learning rule which is described in the next section. A sample Hopfield Network is shown in Fig. 1.

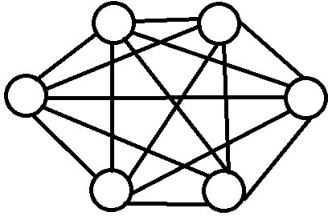

Fig 1. A sample Hopfield Network consisting of 6 neurons.

The Energy of a Hopfield net is described using the equation

$$E = -\frac{1}{2}\left(\sum_{i=1}^{N}\sum_{j=1}^{N} W_{ij}\xi_i^\mu \xi_j^\mu\right) \quad (2)$$

To understand this we try to look at a visualization of the energy landscape which is plotted between the energy vs patterns. Since the patterns are N dimensional vectors this landscape is only a visualization and not the exact graph which is plotted on hypersphere. A visualisation is given in Fig. 2.

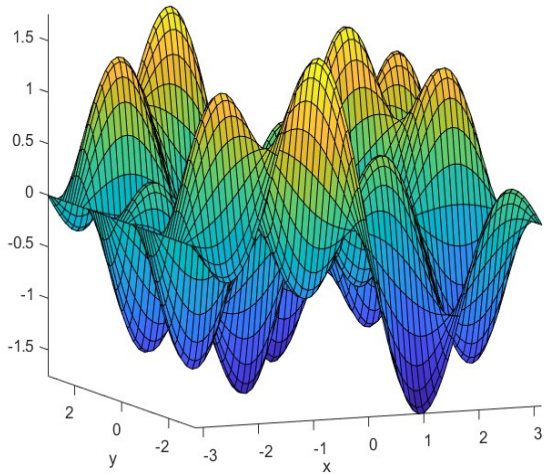

Fig. 2. The 3D energy landscape of a Hopfield network before training.

## II. COMPARISON BETWEEN DNN AND ITERATIVE LEARNING ASSOCIATIVE MODEL NETWORK

### A. Training Process

#### 1) DNN

Training is the process of finding 'weights' of the model which will satisfy the input output mapping. Weights are parameters of the model which are used along with the inputs to produce output. In other words, the output is a function of the inputs as well as the model weights. Before training, these weights may be initialized randomly. During training, batches of inputs are taken and according to the initial random weights. Then, an objective function, as defined by the programmer, is calculated, which is a function of the predicted value as well as the actual value. Then, the weights are updated so that this objective function(or loss) is minimized. This is done through various optimization rules such as Stochastic Gradient Descent(SGD)[9], Adam[10], RMSProp[11] and so on. This process is continued for many iterations until a reasonable accuracy is achieved. We have used SGD and RMS error as our optimization function and objective function respectively.

One important distinction to be made here is that we are not interfering with the inputs of the model during training. We only change the weights of the model as shown in Fig. 3. We can visualize the training process by plotting the objective function against the weights of the model. The weights get updated in each iteration of the training and finally converge to a minima. However, no updates are made to the input patterns during this process. This is not the case for Associative Memory Models, where modifications are made in the patterns.

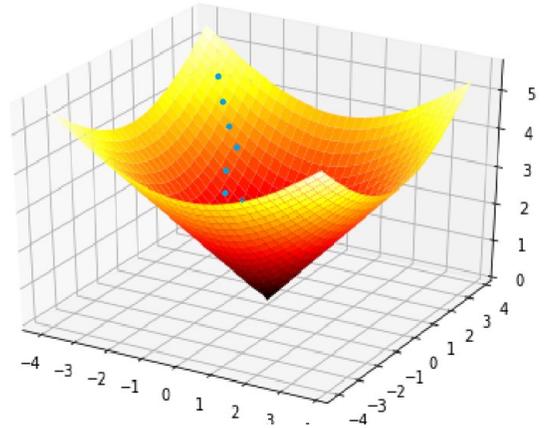

Fig. 3. Gradient Descent, where the weights are updated towards minima. Here, the weights are initialized randomly and as the training proceeds, the path as shown is followed and a minima is reached.

#### 2) Iterative Learning Associative Networks

The training process of Iterative Learning Associative Model Network depends on the learning rule. In the iterative learning rule, the weight matrix is initialized to the Hebbian weight matrix and then a correction value is added in each iteration. The correction is calculated by flipping states of the patterns randomly and checking the lowering of energy since the patterns of the network should be the lowest in energy compared to the vicinity of the pattern in terms of hamming distance. To visualize the change of the weight matrix in terms of the energy landscape, see Fig. 2, Fig. 4 and Fig. 5.

*B. Adversarial Inputs*

Adversarial inputs are slightly modified input patterns which were previously recognized by the model but not after the modification. However, These patterns seem similar to the human eye. [12] first showed the existence of adversarial inputs. The existence of such inputs poses a threat to privacy and security with the increasing application of deep learning models in fields concerning human lives.

*1) DNN*

Deep neural networks are shown to be highly prone to adversarial attacks [13]. After [12] discovered the existence of adversarial inputs, many more such attacks have been invented. A defense mechanism which is robust to all such attacks has not been created to the best of our knowledge. For all the existing deep neural networks, there exist attacks which mislead the model without misleading the human eye.

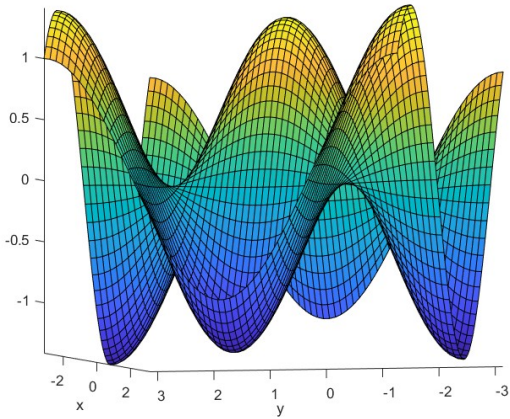

Fig. 4. The 3D energy landscape of a Hopfield network after training

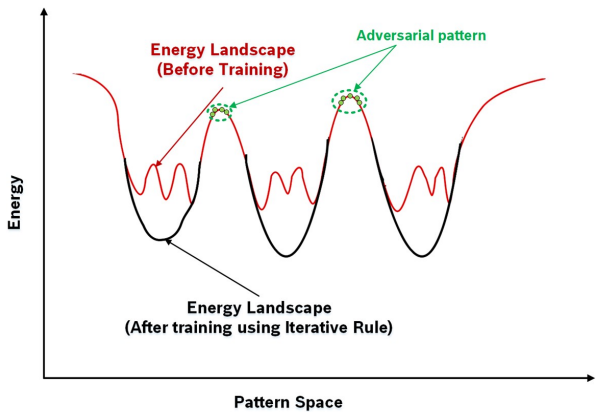

Fig. 5. This figure shows the change in landscape while training in an iterative rule. The black landscape is the trained landscape. The red landscape inside is the previous landscape.

*2) Associative Memory*

An Iterative Learning Associative Model network is not based on the conventional gradient based optimization. Hence, most of the adversarial attacks don't work on it. As an experiment, we generated 40 adversarial images from the MNIST dataset, for a dense network and recalled them using a Iterative Learning Associative Model Network. We found that the number of images which were misclassified by the Iterative Learning Associative Model Network was 0.7 .

This experiment proves the model dependence of adversarial attacks. We define model dependence of adversarial attacks as the property of adversarial attacks being successful on one type of learning rule but not on others. Similarly, model independence is the property of an adversarial attack by which it is successful for any learning rule. However, [14] proved that adversarial attacks can be viewed as model independent, by creating transfer based attacks. Attacks against an ensemble of individual models are generally successful against other models as well. Thus, in case of Deep Neural Networks, adversarial attacks can be transferred from one model to another.

We claim that a similar model independence can always be created by introducing more types of models in such an ensemble.

Thus, the attacks generated against such an ensemble consisting of associative memory models and recurrent models should be model independent in a rough sense. However, strictly speaking, adversarial attacks are always model dependent.

*C. Existence of Replicas in the Weight Space*

*1) DNN*

Usually almost every architecture has a few dense layers before the output layers. Dense layers are important to the model since they do the actual classification from the features extracted by the convolutions layers. According to [15], there exist transformations which when applied to the weight vector of fully connected dense layers, keeps the output unchanged. These transformations are sign flips and weight(belonging to the same layer) interchanges. Thus, multiple weight vectors exist for the same mapping. It has also been proved that these weights take the form of a wedge and cone in the weight space. These cones are called equioutput cones. A visual representation is shown in Fig. 6 and Fig. 8. The paper further explains that the higher the dimensionality of the weight space, the easier it is to converge to a minimum. For more information, we refer readers to [15].

For every set of patterns in the input(pattern-space), we define the weights learned by a model so as to minimize the objective function, as the corresponding weight vector. Thus, we define the relation between weight space and pattern space to be model dependent. However, for each such relation we can make interesting observations.

We define the cone formed in the weight space corresponding to data points in the training set, as the exact cone. A similar cone is formed for all those patterns which are not in the data-set but are predicted/classified correctly by the model. We term this as the generalized cone. These equioutput cones are formed by sign flips and interchanges in weights just as shown in Fig. 6. Using the correlation between weight-space and pattern-space as described above, we get equioutput cones for each of the cones using these transformations(sign flips and interchanges). We verified the existence of equioutput cones by defining an order of ascendance between the weights of a layer in the weight vector viz. we initialized them in a particular order, and empirically viewed that the order is maintained during any time instance of the training. Thus, a weight vector does not change its cone during training. This was true for every permutation of the weights of a layer in the weight vector.

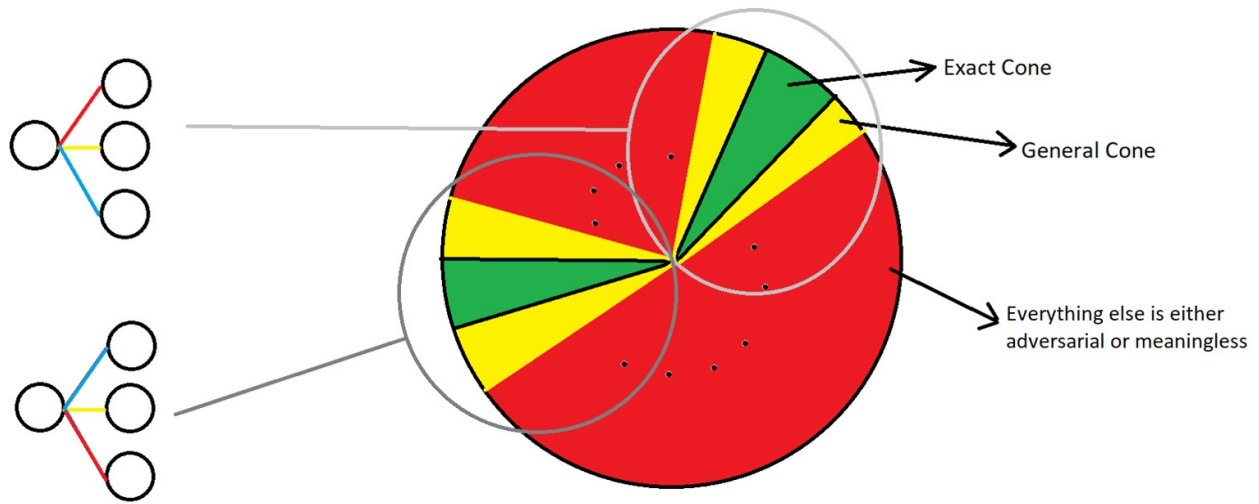

Fig 6. The existence of equioutput cones, visualized, according to [12]. To the left is an example layer of a dense network. Each color represents a weight. Since we have a fully connected net, even on interchanging red and blue weights, the output will be the same. Hence, we get an equioutput cone.

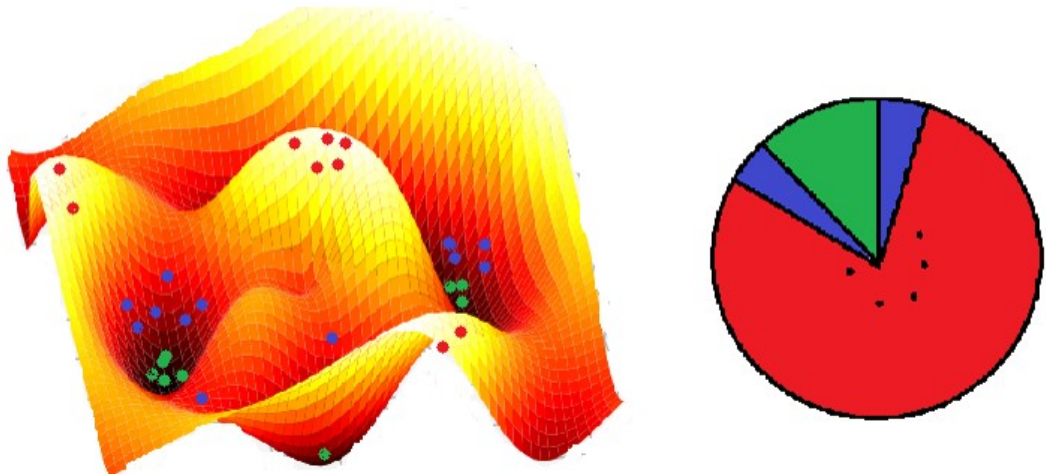

Fig 7. The green points correspond to a minima, mapping to the exact cone(green part in hypersphere), blue points correspond to the generalized cone(blue part in the hypersphere), and the red points correspond to the Non Exact, non generalized cone(red region in the hypersphere). This way, all the points on the objective function belong to one of these cones.

We term the remaining part of the hypersphere as the Non Exact Non Generalized Cone. This corresponds toadversarial inputs as well as irrelevant inputs, which are not images which are formed by making small amount of perturbations to the original image so that they are still classifiable by the human eye, but completely mislead the model. Irrelevant images are all other combinations of the pixels. For example, a completely black image is irrelevant for an MNIST dataset.

A 3D representation of the hypersphere, along with exact and generalized cones, is shown in Fig. 8. We also show an analogy between the position on the objective function and the hypersphere in Fig. 7. The points very near to the minimas correspond to the exact cone. Points a little further away but closer to the minima than the maxima correspond to the generalized cone. All other points correspond to the Non Exact Non Generalized Cone. We believe that the points corresponding to the maxima have some relation with being adversarial. predicted correctly by the model.

We emphasize that we have shown the above results empirically and a mathematical proof will be taken up as future work. However, it is not feasible to separate adversarial patterns from irrelevant patterns. This method relies only on the model's prediction, given an image and so, to classify a pattern as adversarial, human decision is needed. Hence, even for simple datasets like MNIST, which have 784 pixels per image and 70,000 images and only considering binary images, it is humanely impossible to classify $2^{784}$ - 70000 images.

Applying the transformations pointed out in [12] we can say that equioutput cones exist in case of exact, generalized as well as non exact and non generalized cones. Moreover, we claim that the volume of the entire hypersphere of the weight space is filled by these cones, i.e.

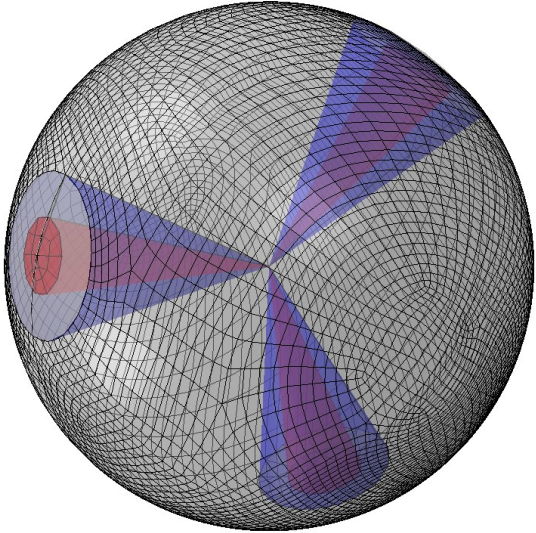

Fig. 8 The red region denotes the exact cone. The blue region, surrounding the exact cone denotes the generalized cone. The cones are made for visualization only. In reality, the cones cover a minuscule volume of the hypersphere.

$P_u = P_t + P_g + P_a + P_r$

And

$W_t = W_e + W_g + W_n$

Where

$P_u$ = Total patterns in the pattern space
$P_t$ = Training patterns
$P_g$ = Generalized patterns, which are predicted correctly but not been trained on, by the model
$P_a$ = Adversarial patterns
$P_r$ = irrelevant patterns
$W_t$ = Total possible weights
$W_e$ = Weights corresponding to the exact cone(corresponding to $P_t$)
$W_g$ = Weights corresponding to the generalized cone(corresponding to $P_g$)
$W_n$ = Weights corresponding to the Non Exact Non Generalized Cone(corresponding to $P_a + P_r$)

### 2) Associative Memory

The equioutput transformation given in [12] are only for deep neural networks. If the same transformation is applied to the Hopfield weight matrix, the same results are not obtained. So there might be some equioutput transformations which are not in mathematically closed forms, but there is no evidence of it.

### III. CONCLUSION AND DISCUSSION

In this paper, we attempted to characterize the hyper sphere of solution space in terms of weights pertaining to exact solution, generalized solutions and weights space belonging to adversarial patterns. We explored two very different learning rules used for a pattern recognition and classification tasks viz. feed-forward deep networks and associative learning rule. We compared different properties like the training procedure, existence of replica weights and robustness against adversarial examples. We see that recalling into a final output image in case of Iterative Learning Associative Model Network is much more advantageous against adversarial attacks. In deep neural networks, the linear layers at the end of the network output scores and are seen to be easily mislead, which is not the case in Iterative Learning Associative Model Neural Networks. Hence, we show the advantage of recall against classification. [16] attempted to combine classification with recall and showed that it indeed is helpful in improving robustness against adversarial attacks. A mathematical proof for proving the existence of replicas in the weight space is left as future work. Also, a definitive relation between the maxima of the graph of the objective function against weights or patterns is yet to be established and is left as future work.

ACKNOWLEDGMENT

Saurav Musunuru and Jay N. Paranjape gratefully acknowledge the opportunity to intern at Robert Bosch and the permission by the Indian Institute of Technology Delhi. The authors are grateful towards K. Srinivasan for helping out with the 3D plots in the paper.